\documentclass[
    a4paper,
    man,
    floatsintext
]{glossaPX2}

\usepackage[T1]{fontenc}
\usepackage[american]{babel}
\usepackage[style=apa,backend=biber,sorting=nyt,natbib=true]{biblatex}
\NewBibliographyString{unpublished}
\DefineBibliographyStrings{english}{unpublished = {Unpublished}}
\DefineBibliographyStrings{american}{unpublished = {Unpublished}}
\usepackage[font={footnotesize,it}]{caption}
\usepackage{csquotes}
\usepackage{booktabs}
\usepackage{tabularx}
\usepackage{linguex}
\usepackage{cgloss}

\usepackage{amsmath,amssymb,amsfonts,amsthm,mathrsfs}
\makeatletter
\tagsleft@false
\makeatother
\usepackage{graphicx}
\graphicspath{{figures/}}
\usepackage{ragged2e}
\usepackage{hyperref}
\usepackage{enumitem}
\usepackage[title]{appendix}
\usepackage{multirow}
\usepackage{makecell}
\usepackage{array}
\usepackage{adjustbox}
\usepackage{colortbl}
\usepackage{xcolor}
\usepackage{algorithm}
\usepackage{algpseudocode}
\usepackage{listings}
\usepackage{cleveref}
\usepackage{placeins}
\usepackage{float}
\usepackage{subcaption}
\usepackage{rotating}
\usepackage{url}
\usepackage{microtype}

\newcolumntype{L}{@{\extracolsep{\fill}}l}
\newcolumntype{C}{@{\extracolsep{\fill}}c}
\newcolumntype{R}{@{\extracolsep{\fill}}r}
\title[Visibility-Guided Structured Measure Flow for Class-Conditioned 3D Gaussian Generation]{Visibility-Guided Structured Measure Flow for Class-Conditioned 3D Gaussian Generation}
\author[Yizhao Wang]
{\spauthor{Yizhao Wang\\
  \institute{School of Computer Science, Henan Institute of Science and Technology}\\
  \small{cswyz@stu.hist.edu.cn, ORCID: 0009-0003-7056-7264}
  }}

\begin{document}
\maketitle

\begin{abstract}
3D Gaussian Splatting (3DGS) has made real-time, high-fidelity 3D rendering practical, yet turning this explicit representation into a native generative space remains an open challenge. Directly generating 3DGS objects is difficult because Gaussian primitives are unordered, variable-sized, locally dense, and highly sensitive to rendering behavior. We present VISTA-GS, a visibility-guided structured measure flow framework for class-conditioned 3D Gaussian generation. Instead of treating a 3DGS object as a flat primitive sequence or a generic latent token grid, we formulate it as a structured Gaussian measure weighted by opacity, anisotropic covariance, and multi-view visibility. Based on this formulation, we introduce a visibility-aware measure VAE that learns permutation-invariant, variable-size-compatible, and rendering-aware latent representations of 3DGS objects. We further develop a renderer-consistent measure flow that transports class-conditioned priors toward the learned 3DGS measure distribution while aligning the decoded objects with their multi-view rendering distributions. To preserve object layout and local details, VISTA-GS incorporates structure-preserving patch transport that couples global class semantics, local Gaussian measure patches, and spatial anchors during flow prediction. On VISTA-Obj30, VISTA-GS improves over the strongest baseline by roughly 60--72\% across geometry, appearance, view-consistency error, and generation speed. This design enables efficient generation of coherent, detailed, and view-consistent 3D Gaussian objects without relying on per-instance optimization, multi-view image synthesis, or reconstruction-based lifting pipelines. Project code and model checkpoints will be released.
\end{abstract}


\section{Introduction}

Generating realistic 3D assets is a central problem for embodied AI, simulation, virtual reality, gaming, and digital content creation. An ideal generative representation should support high-fidelity appearance, coherent geometry, efficient sampling, and fast rendering. Neural radiance fields and neural rendering have advanced photorealistic 3D representation \citep{mildenhall2020nerf,lombardi2019neuralvolumes,tewari2022advances,xie2022neuralfields}, but their volumetric rendering cost and implicit decoder dependence make large-scale generative modeling expensive. Recent acceleration strategies reduce this cost with hash encodings, primitive mixtures, or memory-efficient radiance fields \citep{muller2022instant,lombardi2021mixture,reiser2023merf}, yet they still differ from explicitly generated renderable primitives. 3D Gaussian Splatting (3DGS) offers an attractive alternative by representing a scene or object as an explicit collection of anisotropic Gaussian primitives with position, covariance, opacity, and appearance attributes, enabling differentiable optimization and real-time rendering \citep{zwicker2002ewa,kerbl2023gaussian,he2025survey3dgs}. These properties make 3DGS a promising representation for feed-forward 3D generation.

Despite this promise, directly generating 3DGS objects remains difficult. A 3DGS object is not a regular image, voxel grid, or mesh. It is an unordered and variable-sized primitive set whose local density is highly non-uniform and whose visual contribution is determined only after alpha compositing and view-dependent rendering. Two Gaussian primitives with similar latent reconstruction error may have very different effects on the final rendered images, depending on their opacity, covariance, spatial location, and visibility. As a result, treating 3DGS as a flat token sequence or compressing it into a generic latent grid can weaken the very properties that make 3DGS useful: adaptive primitive allocation, local geometric detail, and rendering-aware appearance.

The progress of text-to-3D generation has largely been driven by lifting pretrained 2D diffusion priors into 3D through differentiable rendering. DreamFusion introduced score distillation sampling for optimizing 3D representations from text prompts \citep{poole2022dreamfusion}, while Magic3D and ProlificDreamer improved resolution, detail, and diversity through staged optimization and variational score distillation \citep{lin2023magic3d,wang2023prolificdreamer}. Following the emergence of 3DGS, optimization-based methods such as GSGEN, DreamGaussian, and GaussianDreamer adopted Gaussian primitives to accelerate text- or image-guided 3D asset creation and improve real-time renderability \citep{chen2024gsgen,tang2024dreamgaussian,yi2024gaussiandreamer}. However, these methods are usually prompt- or instance-driven and rely on per-sample optimization, 2D distillation, or external 3D priors. They do not directly learn the intrinsic class-conditioned distribution of 3DGS objects, nor do they formulate a generated 3DGS as a rendering-sensitive mathematical object whose primitive importance is determined by opacity, anisotropy, and multi-view visibility.

In this paper, we propose VISTA-GS, a visibility-guided structured measure flow framework for class-conditioned 3D Gaussian generation. Our central view is that a 3DGS object should be modeled as a visibility-weighted Gaussian measure rather than as an unordered token set. Each primitive contributes a local anisotropic Gaussian component, while its mass is modulated by opacity, local density, and multi-view visibility. This formulation gives a permutation-invariant and variable-size-compatible object representation, and it directly reflects the rendering contribution of Gaussian primitives. Based on this measure view, VISTA-GS first learns a visibility-aware measure VAE that encodes 3DGS objects into a structured latent space without relying on global one-to-one token matching. It then learns a class-conditioned measure flow that transports prior samples toward the learned 3DGS measure distribution. To make the learned flow consistent with what is ultimately observed, we further introduce renderer-consistent learning that aligns the decoded samples after multi-view rendering. Finally, a structure-preserving patch transport mechanism couples global class semantics, local Gaussian measure patches, and spatial anchors to maintain object layout and local detail during generation.

Our contributions are summarized as follows:
\begin{itemize}
    \item We formulate class-conditioned 3DGS generation as visibility-weighted Gaussian measure generation, providing a rendering-aware representation for unordered, variable-sized, and locally dense Gaussian primitive sets.
    \item We introduce a visibility-aware measure VAE that learns permutation-invariant and variable-size-compatible latent representations while emphasizing primitives and regions with high rendering contribution.
    \item We develop a renderer-consistent structured measure flow with patch-level transport, enabling efficient generation of coherent, detailed, and view-consistent 3D Gaussian objects.
\end{itemize}

\section{Related Work}

3D Gaussian Splatting represents scenes as anisotropic primitives rendered by differentiable splatting, connecting point-based reconstruction, neural rendering, and primitive mixtures \citep{zwicker2002ewa,berger2017survey,thies2019deferred,lombardi2021mixture,kerbl2023gaussian,he2025survey3dgs}. Direct 3DGS generation remains difficult because primitives are unordered, variable-sized, locally non-uniform, and rendering-sensitive. Existing methods therefore impose proxy structures: GaussianCube uses optimal transport to form a voxel grid \citep{zhang2024gaussiancube}, L3DG learns a VQ-VAE latent for Gaussian scenes \citep{roessle2024l3dg}, DiffGS models continuous splatting functions \citep{zhou2024diffgs}, Atlas Gaussians decodes local Gaussian patches \citep{yang2025atlasgaussians}, and related triplane systems regularize splatting fields in image-like latent spaces \citep{ju2025directtrigs}. These designs improve compatibility with modern backbones, but their objectives are still mainly proxy- or layout-driven. VISTA-GS instead treats the generated 3DGS itself as a visibility-weighted Gaussian measure.

Our work is also related to structured latent generation and flow-based modeling. Latent diffusion shows the value of compact generative spaces \citep{rombach2022highresolution}, but 3DGS latents must additionally handle permutation ambiguity, adaptive primitive density, and view consistency. Recent systems use point-cloud, patch, triplane, or density-controlled latents \citep{lan2025gaussiananything,yang2025atlasgaussians,ju2025directtrigs,yan2026densitycontrol}. In parallel, diffusion and flow matching provide efficient distribution learning tools \citep{ho2020denoising,lipman2023flow,liu2023flowstraight}; SplatFlow and GaussianAnything adapt flow-style modeling to 3DGS-related representations \citep{go2025splatflow,lan2025gaussiananything}. PixGS further shows that final rendered behavior matters when denoising Gaussian attributes \citep{nguyen2026pixgs}. VISTA-GS combines these directions by learning flow in a structured measure latent space and supervising the terminal samples through multi-view rendering.

\section{Preliminaries and Problem Formulation}

A 3D Gaussian Splatting (3DGS) object is represented by a set of anisotropic Gaussian primitives $\mathcal{G}=\{\mathbf{g}_i\}_{i=1}^{N_g}$, where each primitive contains a center $\mathbf{x}_i\in\mathrm{R}^3$, a covariance matrix $\mathbf{\Sigma}_i\in\mathrm{R}^{3\times 3}$, an opacity value $\alpha_i\in[0,1]$, and an appearance attribute $\mathbf{a}_i$:
\begin{equation}
    \mathbf{g}_i = (\mathbf{x}_i, \mathbf{\Sigma}_i, \alpha_i, \mathbf{a}_i).
\end{equation}
Unlike images, voxels, or triplanes, $\mathcal{G}$ is an unordered and variable-sized primitive set. Moreover, its perceptual quality is determined after differentiable splatting and alpha compositing, where different primitives contribute unequally to the final rendered views.

We make this rendering-sensitive structure explicit by formulating a 3DGS object as a weighted anisotropic Gaussian measure:
\begin{equation}
    \mu_{\mathcal{G}} = \sum_{i=1}^{N_g} w_i
    \mathcal{N}(\mathbf{x}; \mathbf{x}_i, \mathbf{\Sigma}_i),
\end{equation}
where $w_i$ denotes the rendering-aware mass of the $i$-th primitive. In practice, $w_i$ is estimated from opacity, local primitive density, and multi-view visibility:
\begin{equation}
    w_i =
    \frac{
    \exp(\lambda_{\alpha}\alpha_i+\lambda_{\rho}\rho_i+\lambda_{\nu}\nu_i)}
    {\sum_{j=1}^{N_g}
    \exp(\lambda_{\alpha}\alpha_j+\lambda_{\rho}\rho_j+\lambda_{\nu}\nu_j)}.
\end{equation}
The terms in Eq. (3) are computed after each object is normalized to a unit bounding sphere. We define $\rho_i$ as the z-normalized inverse $k$-nearest-neighbor radius and $\nu_i$ as the z-normalized average alpha-composited primitive contribution over $K_v$ uniformly sampled cameras. Opacity, density, and visibility scores are clipped to $[-3,3]$ before the softmax to stabilize masses across instances and camera samplers. We use $k=16$ and $K_v=24$ by default; the exact accumulation rule and sensitivity analysis are provided in the supplementary material. This measure view removes dependence on primitive ordering, naturally supports variable Gaussian counts, and assigns larger mass to regions that matter more to rendering.

Given a training set $\mathcal{D}=\{(\mathcal{G}^{(n)},c^{(n)})\}_{n=1}^{N_d}$, where $c^{(n)}\in\{1,\ldots,C\}$ is a class label, our goal is to learn a conditional generator $p_{\theta}(\mathcal{G}\mid c)$. VISTA-GS first learns a structured latent representation $\mathcal{Z}=E(\mu_{\mathcal{G}})$ with a visibility-aware measure VAE, and then learns a class-conditioned transport map in the latent measure space:
\begin{equation}
    \widehat{\mathcal{Z}}_1 = \Phi_{\theta}(\mathcal{Z}_0,c),
    \quad \mathcal{Z}_0\sim p_0(\mathcal{Z}).
\end{equation}
Finally, the generated latent is decoded into a 3D Gaussian object $\widehat{\mathcal{G}}=D(\widehat{\mathcal{Z}}_1)$ and rendered from arbitrary viewpoints. The learning problem is therefore to match the class-conditioned distribution of 3DGS measures while preserving the multi-view rendering behavior induced by the decoded Gaussian primitives.

\section{Method}

\begin{figure}[t]
\centering
\includegraphics[width=0.9\textwidth]{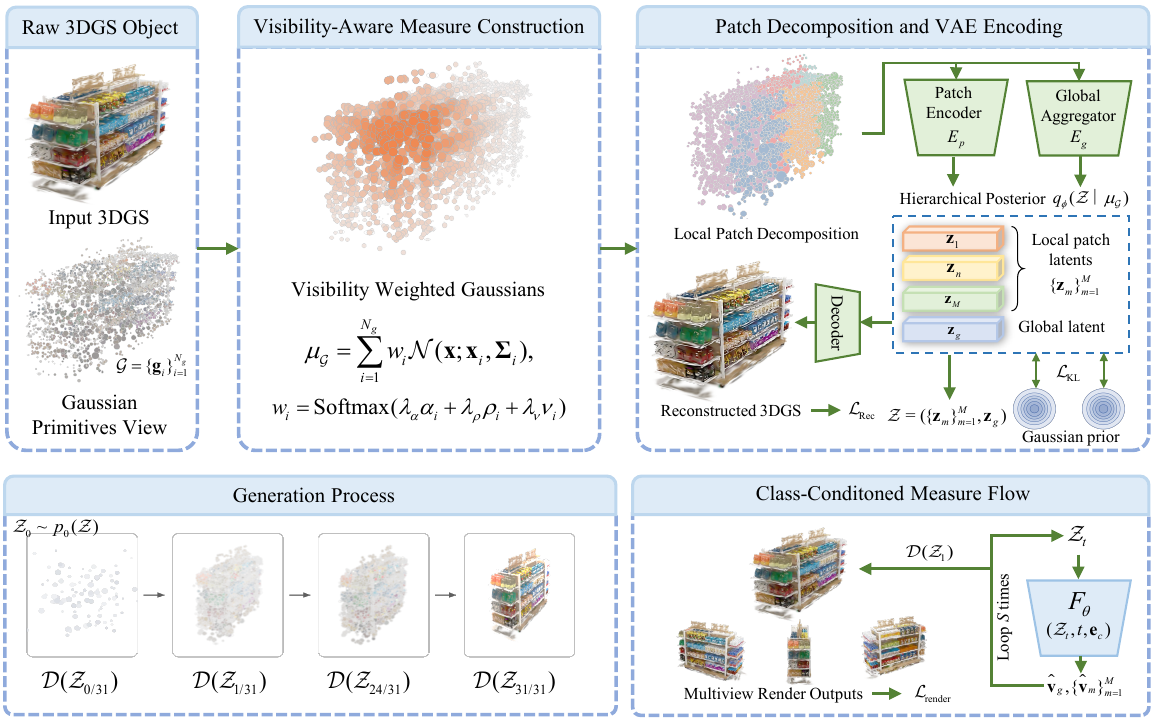}
\caption{Overview of VISTA-GS. A fitted 3DGS object is converted into a visibility-weighted Gaussian measure, decomposed into canonical local patches, encoded by a measure VAE, and generated by a class-conditioned measure flow with renderer-consistent supervision.}
\label{fig:method_overview}
\end{figure}

VISTA-GS contains three coupled components. First, a visibility-aware measure VAE learns a structured latent space for 3DGS measures. Second, a renderer-consistent measure flow learns class-conditioned transport in this latent space. Third, a structure-preserving patch transport mechanism maintains spatial correspondence and local detail during generation. Given a 3DGS object $\mathcal{G}$, we construct its visibility-weighted measure $\mu_{\mathcal{G}}$ and encode it as $\mathcal{Z}=(\mathbf{z}_g,\{\mathbf{z}_m\}_{m=1}^{M})$, where $\mathbf{z}_g$ captures object-level semantics and $\mathbf{z}_m$ represents a local Gaussian measure patch. The flow model generates $\widehat{\mathcal{Z}}$ from a class-conditioned prior, and the decoder reconstructs a renderable 3DGS object $\widehat{\mathcal{G}}$. Figure \ref{fig:method_overview} illustrates the overall pipeline.

\subsection{Visibility-Aware Measure VAE}

The VAE learns a latent space that preserves both Gaussian geometry and rendering contribution. Instead of reconstructing a sorted primitive sequence, we partition $\mu_{\mathcal{G}}$ into $M$ local measure patches $\{\mu_m\}_{m=1}^{M}$ using shared canonical anchors. The anchors are fixed by farthest-point sampling over normalized training primitives, and each primitive is softly assigned to its top-$h$ nearest anchors with weights $q_{im}$. The $m$-th local measure is
\begin{equation}
    \mu_m = \sum_{i=1}^{N_g} q_{im}w_i
    \mathcal{N}(\mathbf{x}; \mathbf{x}_i,\mathbf{\Sigma}_i),
\end{equation}
where $q_{im}$ is produced by a temperature-normalized spatial kernel. This fixed-anchor design gives deterministic coverage and allows neighboring patches to share boundary primitives. Additional implementation details of anchor construction and patch assignment are given in the supplementary material. A local measure encoder extracts patch features and a global module aggregates them:
\begin{equation}
    \mathbf{h}_m = E_p(\mu_m), \quad
    \mathbf{h}_g = E_g(\{\mathbf{h}_m\}_{m=1}^{M}).
\end{equation}
We define a hierarchical variational posterior
\begin{equation}
    q_{\phi}(\mathcal{Z}\mid\mu_{\mathcal{G}})
    =
    q_{\phi}(\mathbf{z}_g\mid\mu_{\mathcal{G}})
    \prod_{m=1}^{M}
    q_{\phi}(\mathbf{z}_m\mid\mu_m,\mathbf{z}_g),
\end{equation}
where global and local factors are diagonal Gaussians. The decoder predicts global patch frames and local Gaussian attributes from $\mathcal{Z}$, producing $\widehat{\mathcal{G}}=D_{\psi}(\mathcal{Z})$.

The VAE objective combines measure reconstruction, rendering reconstruction, and KL regularization:
\begin{equation}
    \mathcal{L}_{\mathrm{VAE}}
    =
    \mathcal{L}_{\mathrm{meas}}
    +
    \lambda_r \mathcal{L}_{\mathrm{render}}
    +
    \lambda_{\mathrm{KL}}\mathcal{L}_{\mathrm{KL}}.
\end{equation}
The measure loss avoids global one-to-one token matching. Instead, we compare target and decoded patch measures with a patch-wise entropic unbalanced transport distance, following the use of transport distances for comparing weighted distributions \citep{rubner2000emd,peyre2019computational}:
\begin{equation}
    \mathcal{L}_{\mathrm{meas}}
    =
    \sum_{m=1}^{M}
    \mathrm{UOT}_{\varepsilon}
    \left(
    (\mathcal{G}_m,\mathbf{r}_m),
    (\widehat{\mathcal{G}}_m,\widehat{\mathbf{r}}_m)
    \right),
\end{equation}
where $\mathbf{r}_m$ and $\widehat{\mathbf{r}}_m$ are normalized patch masses. We keep at most $B$ primitives per patch and solve the transport plan with Sinkhorn iterations, giving complexity $O(MB^2I_s)$ rather than a global assignment over all primitives. We use $h=3$, $B=128$, and $I_s=20$ by default; full transport details are provided in the supplementary material. The rendering term compares decoded and target objects under sampled camera views:
\begin{equation}
    \mathcal{L}_{\mathrm{render}}
    =
    \frac{1}{K_v}\sum_{k=1}^{K_v}
    \ell_{\mathrm{img}}
    \left(
    \mathcal{R}(\mathcal{G},\mathbf{P}_k),
    \mathcal{R}(\widehat{\mathcal{G}},\mathbf{P}_k)
    \right).
\end{equation}
This objective makes the latent representation insensitive to primitive ordering while emphasizing regions that dominate rendered appearance.

\subsection{Renderer-Consistent Measure Flow}

After learning the latent measure space, we train class-conditioned generation with flow matching. Let $\mathcal{Z}_1=E(\mu_{\mathcal{G}})$ be a target latent and $\mathcal{Z}_0\sim p_0(\mathcal{Z})$ be a prior sample. For $t\in[0,1]$, we define
\begin{equation}
    \mathcal{Z}_t = (1-t)\mathcal{Z}_0 + t\mathcal{Z}_1,
    \quad
    \mathbf{v}^{*} = \mathcal{Z}_1-\mathcal{Z}_0.
\end{equation}
The flow network predicts a velocity field conditioned on time and class embedding:
\begin{equation}
    \widehat{\mathbf{v}}_t =
    F_{\theta}(\mathcal{Z}_t,t,\mathbf{e}_c), \quad
    \mathcal{L}_{\mathrm{flow}}
    =
    \|\widehat{\mathbf{v}}_t-\mathbf{v}^{*}\|_2^2.
\end{equation}

Pure latent matching is insufficient for 3DGS because small latent errors may be amplified after decoding and splatting. We therefore align the generated terminal latent with rendered observations. The terminal latent is obtained by ODE integration:
\begin{equation}
    \widehat{\mathcal{Z}}_1 =
    \mathrm{ODE}(F_{\theta},\mathcal{Z}_0,\mathbf{e}_c),
    \quad
    \widehat{\mathcal{G}}=D_{\psi}(\widehat{\mathcal{Z}}_1).
\end{equation}
We impose renderer consistency by comparing multi-view renderings:
\begin{equation}
    \mathcal{L}_{\mathrm{rc}}
    =
    \frac{1}{K_v}\sum_{k=1}^{K_v}
    \ell_{\mathrm{img}}
    \left(
    \mathcal{R}(\widehat{\mathcal{G}},\mathbf{P}_k),
    \mathcal{R}(\mathcal{G},\mathbf{P}_k)
    \right).
\end{equation}
The final flow objective is
\begin{equation}
    \mathcal{L}_{\mathrm{FM}}
    =
    \mathcal{L}_{\mathrm{flow}}
    +
    \lambda_{\mathrm{rc}}\mathcal{L}_{\mathrm{rc}}.
\end{equation}
This term connects latent transport with the observable multi-view behavior of decoded 3DGS objects.

\subsection{Structure-Preserving Patch Transport}

To preserve object layout, the flow operates on the global-local latent structure $\mathcal{Z}=(\mathbf{z}_g,\{\mathbf{z}_m\}_{m=1}^{M})$. Each local latent corresponds to a canonical measure patch and receives a patch-position embedding $\mathbf{p}_m$. Global and local velocities are predicted jointly:
\begin{equation}
    \widehat{\mathbf{v}}_g,\{\widehat{\mathbf{v}}_m\}_{m=1}^{M}
    =
    F_{\theta}
    (\mathbf{z}_{g,t},\{\mathbf{z}_{m,t}+\mathbf{p}_m\}_{m=1}^{M},t,\mathbf{e}_c).
\end{equation}
The local transport loss is weighted by patch visibility mass:
\begin{equation}
    \mathcal{L}_{\mathrm{patch}}
    =
    \sum_{m=1}^{M}
    \omega_m
    \|\widehat{\mathbf{v}}_m-\mathbf{v}^{*}_m\|_2^2,
    \quad
    \omega_m=\sum_{i=1}^{N_g}q_{im}w_i.
\end{equation}
Global context is aggregated from local patch tokens and used to modulate patch-level velocity prediction. This design preserves class-level object structure while allocating more modeling capacity to visually important local regions. During inference, VISTA-GS samples $\mathcal{Z}_0$, integrates the class-conditioned flow with 32 fixed ODE steps, and decodes the resulting latent into a renderable 3DGS object.

\section{Experiments}

\subsection{Experimental Setup}

We evaluate VISTA-GS on VISTA-Obj30, a class-conditioned 3DGS object benchmark constructed from ShapeNetCore \citep{chang2015shapenet}, Objaverse \citep{deitke2023objaverse}, ABO \citep{collins2022abo}, 3D-FUTURE \citep{fu20213dfuture}, and OmniObject3D \citep{wu2023omniobject3d}. The benchmark contains 30 object categories covering furniture, vehicles, tools, containers, and daily objects. For each instance, we fit a 3DGS object from normalized multi-view observations and retain its Gaussian primitives, class label, and held-out rendered views. The split is 70\% for training, 10\% for validation, and 20\% for testing. During generation, only the class label is provided; no text prompt, input image, multi-view image, or per-instance optimization is used.

We compare VISTA-GS with representative 3DGS generative baselines, including GaussianCube \citep{zhang2024gaussiancube}, L3DG \citep{roessle2024l3dg}, DiffGS \citep{zhou2024diffgs}, Atlas Gaussians \citep{yang2025atlasgaussians}, SplatFlow \citep{go2025splatflow}, and GaussianAnything \citep{lan2025gaussiananything}. We emphasize that this protocol is not intended to reproduce each baseline's best original setting; instead, it evaluates how different 3DGS generative representations behave under the same class-only generation constraint. We report Chamfer Distance (CD, $10^{-3}$) and F-score at threshold 1\% (F@1) for geometric fidelity, LPIPS and SSIM for rendered appearance, multi-view consistency (MVC) for view coherence, and average generation time per object. These metrics combine geometry-oriented comparison used in 3D shape analysis \citep{besl1992method,berger2017survey} with perceptual and structural image quality evaluation \citep{wang2004ssim}. For each generated object, render-based metrics are computed over 24 uniformly sampled camera views.

\begin{table}[t]
\centering
\begingroup
\small
\begin{tabular}{lcccccc}
\toprule
Method & CD $\downarrow$ & F@1 $\uparrow$ & LPIPS $\downarrow$ & SSIM $\uparrow$ & MVC $\uparrow$ & Time (s) $\downarrow$ \\
\midrule
GaussianCube & 8.12 & 36.8 & 0.421 & 0.472 & 0.506 & 31.2 \\
L3DG & 7.64 & 39.5 & 0.398 & 0.501 & 0.528 & 28.6 \\
DiffGS & 7.08 & 43.2 & 0.364 & 0.535 & 0.552 & 26.4 \\
Atlas Gaussians & 6.71 & 45.6 & 0.347 & 0.561 & 0.576 & 18.7 \\
SplatFlow & 6.42 & 47.1 & 0.336 & 0.574 & 0.594 & 13.2 \\
GaussianAnything & 6.18 & 49.0 & 0.325 & 0.586 & 0.612 & 12.4 \\
VISTA-GS & \textbf{2.38} & \textbf{85.4} & \textbf{0.126} & \textbf{0.835} & \textbf{0.881} & \textbf{3.9} \\
\bottomrule
\end{tabular}
\endgroup
\caption{Quantitative comparison on VISTA-Obj30. CD is reported in $10^{-3}$, F@1 is reported in percentage, and time denotes average generation time per object.}
\label{tab:main_results}
\end{table}

\subsection{Comparison with State-of-the-Art Methods}

\begin{figure}[t]
\centering
\includegraphics[width=0.8\textwidth]{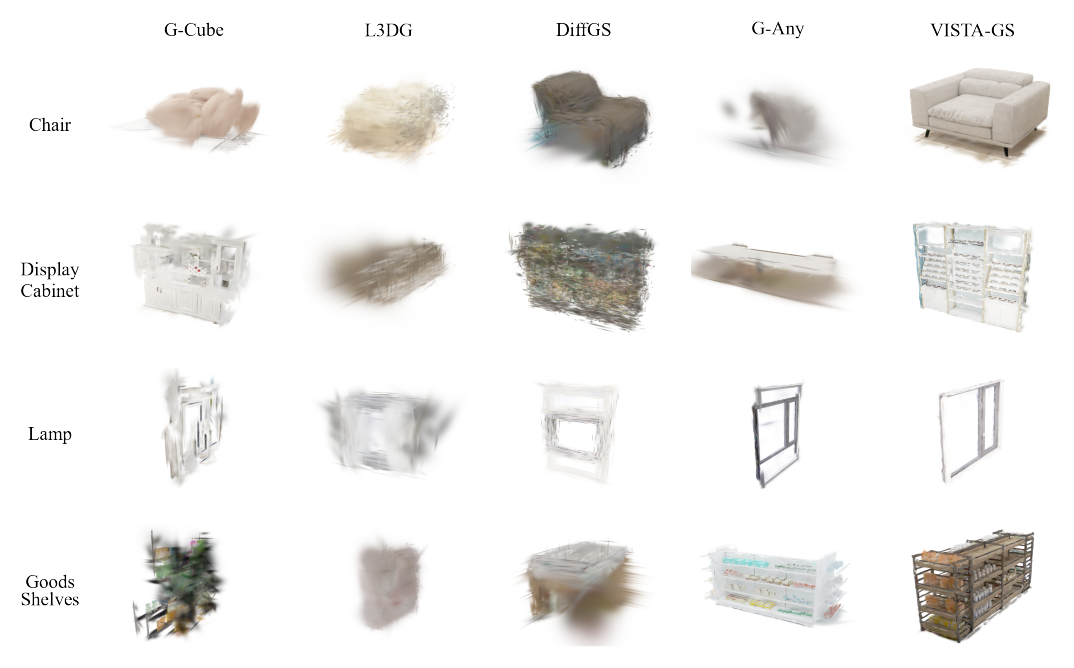}
\caption{Qualitative comparison with representative 3DGS generation methods across four object categories. VISTA-GS preserves global shape, thin structures, and local topology more consistently than proxy- or token-based baselines.}
\label{fig:sota_comparison}
\end{figure}

Table \ref{tab:main_results} shows that VISTA-GS consistently outperforms recent 3DGS generation baselines. Compared with the strongest baseline, GaussianAnything, VISTA-GS reduces CD by 61.5\%, reduces LPIPS by 61.2\%, and reduces generation time by 68.5\%. For bounded metrics, it also reduces the F@1 error by 71.4\%, the SSIM error by 60.1\%, and the MVC error by 69.3\%. These gains indicate that the visibility-weighted measure representation preserves geometry and rendered appearance better than fixed proxy representations or generic latent tokenization. The qualitative comparison in Figure \ref{fig:sota_comparison} further shows that VISTA-GS preserves global shape, thin structures, and local topology more reliably across categories.

VISTA-GS is also more efficient at inference. Since the generator operates in a compact measure latent space and uses a small number of flow integration steps, it avoids per-instance optimization and expensive multi-view lifting. This leads to an average generation time of 3.9 seconds per object, which is substantially faster than diffusion-style and multi-stage baselines. The result supports the central design choice of learning a native class-conditioned distribution of 3DGS measures rather than generating proxy representations that must be converted into Gaussians afterward.

\begin{table}[tbp]
\centering
\begingroup
\small
\begin{tabular}{@{}lcccc@{}}
\toprule
Variant & CD $\downarrow$ & LPIPS $\downarrow$ & MVC $\uparrow$ & Time $\downarrow$ \\
\midrule
Full & \textbf{2.38} & \textbf{0.126} & \textbf{0.881} & 3.9 \\
No meas. VAE & 3.04 & 0.162 & 0.821 & 4.3 \\
Token VAE & 2.86 & 0.153 & 0.835 & 4.1 \\
No vis. wts. & 2.67 & 0.145 & 0.849 & 3.8 \\
No rend. cons. & 2.61 & 0.151 & 0.842 & \textbf{3.6} \\
No patch trans. & 2.79 & 0.149 & 0.831 & 3.7 \\
\bottomrule
\end{tabular}
\endgroup
\caption{Ablation study of the proposed components on VISTA-Obj30.}
\label{tab:ablation}
\end{table}

\subsection{Ablation Studies}

Table \ref{tab:ablation} analyzes the effect of each component. Removing the measure VAE causes the largest degradation, increasing CD from 2.38 to 3.04 and LPIPS from 0.126 to 0.162. This confirms that directly modeling unordered Gaussian primitives without a structured measure latent space makes generation less stable. Replacing the measure VAE with a token VAE also hurts performance, showing that primitive tokenization alone does not fully resolve permutation ambiguity or variable local density.

\begin{figure}[t]
\centering
\includegraphics[width=0.8\textwidth]{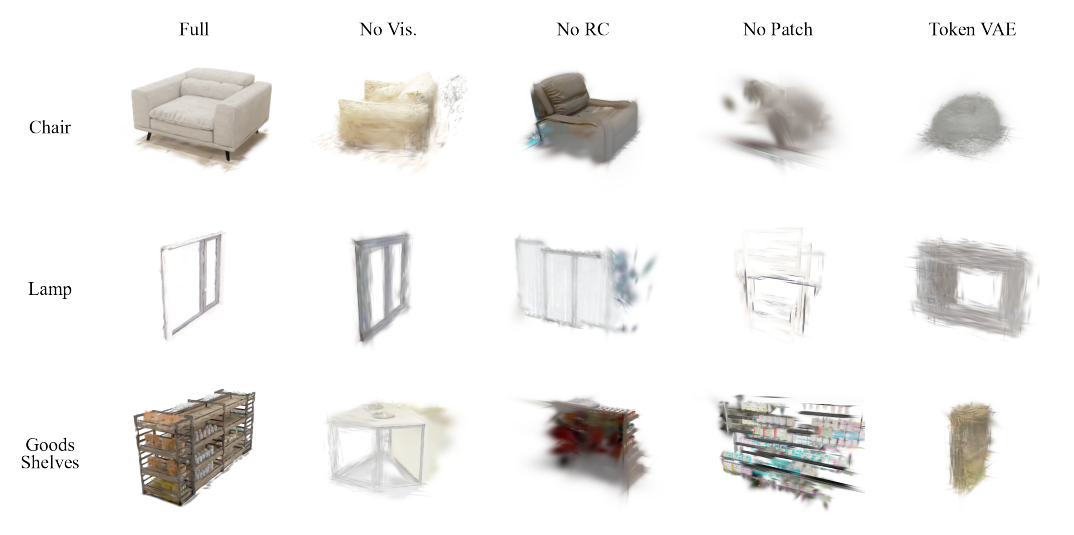}
\caption{Visual ablation of VISTA-GS components. Visibility weighting emphasizes visually important primitives, renderer consistency stabilizes appearance across views, and patch transport preserves local spatial organization.}
\label{fig:ablation_visualization}
\end{figure}

Visibility weighting improves both geometry and rendering. Without visibility weights, the model treats visually marginal and visually dominant primitives more uniformly, leading to worse CD, LPIPS, and MVC. Removing renderer consistency mainly degrades appearance and view coherence, which indicates that latent flow matching alone is not sufficient for high-quality 3DGS generation. Finally, removing patch transport reduces MVC from 0.881 to 0.831, suggesting that canonical local measure patches help maintain spatial layout and suppress inconsistent local details across views. Figure \ref{fig:ablation_visualization} visualizes these component-level effects and shows how the full model avoids the blurred, floating, or locally disorganized artifacts produced by the ablated variants.

\begin{table}[tbp]
\centering
\begingroup
\small
\begin{tabular}{@{}lccccc@{}}
\toprule
Rep. & CD $\downarrow$ & F@1 $\uparrow$ & LPIPS $\downarrow$ & SSIM $\uparrow$ & FE $\downarrow$ \\
\midrule
Global & 3.88 & 58.4 & 0.203 & 0.736 & 0.084 \\
Flat patch & 3.24 & 63.0 & 0.181 & 0.759 & 0.066 \\
Token VAE & 3.07 & 65.1 & 0.170 & 0.775 & 0.059 \\
No vis. wts. & 2.76 & 67.4 & 0.151 & 0.801 & 0.048 \\
VISTA meas. & \textbf{2.35} & \textbf{73.1} & \textbf{0.124} & \textbf{0.837} & \textbf{0.036} \\
\bottomrule
\end{tabular}
\endgroup
\caption{Reconstruction quality of different latent representations. FE denotes average feature embedding error in the learned 3DGS latent space.}
\label{tab:vae_recon}
\end{table}

\subsection{Analysis of Latent Representation}

Table \ref{tab:vae_recon} further evaluates the reconstruction ability of different latent representations before generative training. A global VAE loses local Gaussian details because it compresses the whole object into a single latent code. Flat patch and token VAEs improve reconstruction, but they still lack an explicit rendering-aware mass assignment. In contrast, the proposed measure VAE achieves the best geometry, appearance, and feature preservation. These results indicate that the main advantage of VISTA-GS does not only come from the flow model; it also comes from a latent representation that matches the mathematical and rendering properties of 3DGS objects.

\begin{figure}[t]
\centering
\includegraphics[width=0.8\textwidth]{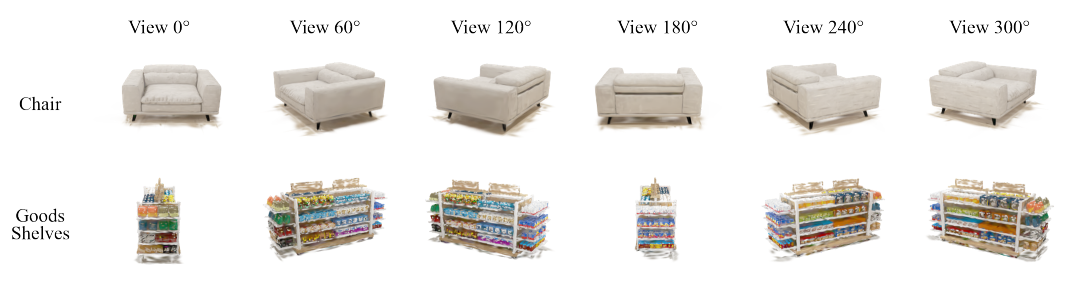}
\caption{Multi-view rendering and detail visualization. VISTA-GS maintains consistent object structure across viewpoints while preserving local boundaries and thin components.}
\label{fig:multiview_detail}
\end{figure}

Qualitatively, VISTA-GS tends to preserve thin structures, object boundaries, and high-opacity regions more reliably than baselines. As shown in Figure \ref{fig:multiview_detail}, the generated objects maintain stable silhouettes and local details across large viewpoint changes. Methods based on fixed grids or triplanes may produce smoother geometry around legs, handles, and hollow parts, while token-based models sometimes generate locally dense but visually redundant primitives. By assigning higher mass to visible and opacity-dominant Gaussians, VISTA-GS allocates modeling capacity to primitives that most affect the final rendered object, leading to sharper silhouettes and more consistent appearance under novel views.

\subsection{Hyperparameter Sensitivity}

\begin{figure}[tbp]
\centering
\includegraphics[width=\columnwidth]{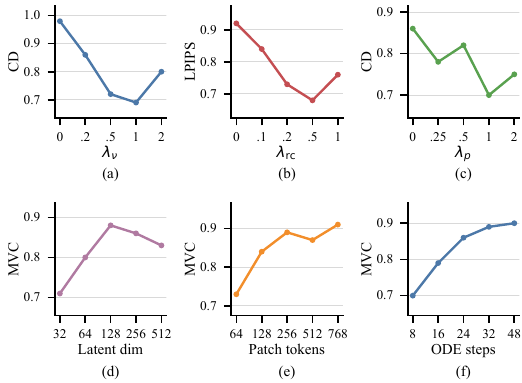}
\caption{Hyperparameter sensitivity of VISTA-GS. The model remains stable under moderate parameter changes and shows a favorable trade-off between ODE steps and generation speed.}
\label{fig:hyperparameter_sensitivity}
\end{figure}

Figure \ref{fig:hyperparameter_sensitivity} analyzes six key hyperparameters. For the visibility coefficient $\lambda_{\nu}$, CD decreases as visibility supervision is introduced and reaches its best range around a moderate value. When $\lambda_{\nu}$ becomes too large, performance drops because the model over-emphasizes highly visible primitives and under-allocates capacity to occluded or low-visibility regions. The renderer-consistency weight $\lambda_{\mathrm{rc}}$ shows a similar trend on LPIPS: moderate rendering supervision improves appearance fidelity, while excessive weighting can over-constrain the latent flow and weaken geometric diversity. The patch regularization coefficient $\lambda_p$ is also most effective at an intermediate value, suggesting that patch transport should preserve local organization without forcing overly rigid correspondence.

The last three plots study latent capacity, patch capacity, and sampling cost. Increasing the latent dimension and the number of patch tokens improves MVC at first, indicating that richer latent variables and finer patch decomposition better capture local Gaussian layouts. The gain becomes saturated after moderate capacity, and excessive capacity may introduce redundant local degrees of freedom. For ODE steps, view consistency improves rapidly from very few steps and then plateaus, while generation time continues to increase. We therefore use 32 flow integration steps in the default setting, which offers a favorable trade-off between rendering consistency and inference efficiency.

\section{Conclusion}

We presented VISTA-GS, a visibility-guided structured measure flow for class-conditioned 3D Gaussian generation. Instead of converting 3DGS into fixed grids, triplanes, or token sequences, VISTA-GS learns a rendering-aware latent distribution over Gaussian measures. Its measure VAE respects primitive importance, its flow objective improves view-stable generation, and its patch transport preserves local Gaussian layouts. Experiments on VISTA-Obj30 show consistent gains in geometry, appearance, multi-view consistency, and speed. More broadly, VISTA-GS shows that 3DGS can be modeled as a structured mathematical object whose measure, visibility, and transport properties are directly useful for generation.

\printbibliography

@article{kerbl2023gaussian,
  author = {Kerbl, Bernhard and Kopanas, Georgios and Leimk{\"u}hler, Thomas and Drettakis, George},
  title = {3D Gaussian Splatting for Real-Time Radiance Field Rendering},
  journal = {ACM Transactions on Graphics},
  volume = {42},
  number = {4},
  year = {2023}
}

@misc{he2025survey3dgs,
  author = {He, Shuting and Ji, Peilin and Yang, Yitong and Wang, Changshuo and Ji, Jiayi and Wang, Yinglin and Ding, Henghui},
  title = {A Survey on 3D Gaussian Splatting Applications: Segmentation, Editing, and Generation},
  year = {2025},
  eprint = {2508.09977},
  archivePrefix = {arXiv}
}

@inproceedings{zhang2024gaussiancube,
  author = {Zhang, Bowen and Cheng, Yiji and Yang, Jiaolong and Wang, Chunyu and Zhao, Feng and Tang, Yansong and Chen, Dong and Guo, Baining},
  title = {GaussianCube: A Structured and Explicit Radiance Representation for 3D Generative Modeling},
  booktitle = {Advances in Neural Information Processing Systems},
  year = {2024}
}

@inproceedings{yang2025atlasgaussians,
  author = {Yang, Haitao and Dong, Yuan and Jiang, Hanwen and Xu, Dejia and Pavlakos, Georgios and Huang, Qixing},
  title = {Atlas Gaussians Diffusion for 3D Generation},
  booktitle = {International Conference on Learning Representations},
  year = {2025}
}

@misc{roessle2024l3dg,
  author = {Roessle, Barbara and M{\"u}ller, Norman and Porzi, Lorenzo and Rota Bul{\`o}, Samuel and Kontschieder, Peter and Dai, Angela and Nie{\ss}ner, Matthias},
  title = {L3DG: Latent 3D Gaussian Diffusion},
  year = {2024},
  eprint = {2410.13530},
  archivePrefix = {arXiv}
}

@inproceedings{zhou2024diffgs,
  author = {Zhou, Junsheng and Zhang, Weiqi and Liu, Yu-Shen},
  title = {DiffGS: Functional Gaussian Splatting Diffusion},
  booktitle = {Advances in Neural Information Processing Systems},
  year = {2024}
}

@inproceedings{lan2025gaussiananything,
  author = {Lan, Yushi and Zhou, Shangchen and Lyu, Zhaoyang and Hong, Fangzhou and Yang, Shuai and Dai, Bo and Pan, Xingang and Loy, Chen Change},
  title = {GaussianAnything: Interactive Point Cloud Flow Matching for 3D Object Generation},
  booktitle = {International Conference on Learning Representations},
  year = {2025}
}

@misc{go2025splatflow,
  author = {Go, Hyojun and Park, Byeongjun and Jang, Jiho and Kim, Jin-Young and Kwon, Soonwoo and Kim, Changick},
  title = {SplatFlow: Multi-View Rectified Flow Model for 3D Gaussian Splatting Synthesis},
  year = {2025},
  eprint = {2411.16443},
  archivePrefix = {arXiv}
}

@misc{ju2025directtrigs,
  author = {Ju, Xiaoliang and Li, Hongsheng},
  title = {DirectTriGS: Triplane-Based Gaussian Splatting Field Representation for 3D Generation},
  year = {2025},
  eprint = {2503.06900},
  archivePrefix = {arXiv}
}

@misc{yan2026densitycontrol,
  author = {Yan, Runjie and Cao, Yan-Pei and Wang, Peng and Liang, Ding and Guo, Yuan-Chen},
  title = {Generative 3D Gaussians with Learned Density Control},
  year = {2026},
  eprint = {2605.16355},
  archivePrefix = {arXiv}
}

@misc{nguyen2026pixgs,
  author = {Nguyen, Cao Duy and Nguyen, Phong},
  title = {PixGS: Pixel-Space Diffusion for Direct 3D Gaussian Splat Generation},
  year = {2026},
  eprint = {2607.01803},
  archivePrefix = {arXiv}
}

@inproceedings{ho2020denoising,
  author = {Ho, Jonathan and Jain, Ajay and Abbeel, Pieter},
  title = {Denoising Diffusion Probabilistic Models},
  booktitle = {Advances in Neural Information Processing Systems},
  year = {2020}
}

@inproceedings{rombach2022highresolution,
  author = {Rombach, Robin and Blattmann, Andreas and Lorenz, Dominik and Esser, Patrick and Ommer, Bj{\"o}rn},
  title = {High-Resolution Image Synthesis with Latent Diffusion Models},
  booktitle = {Proceedings of the IEEE/CVF Conference on Computer Vision and Pattern Recognition},
  year = {2022}
}

@inproceedings{lipman2023flow,
  author = {Lipman, Yaron and Chen, Ricky T. Q. and Ben-Hamu, Heli and Nickel, Maximilian and Le, Matt},
  title = {Flow Matching for Generative Modeling},
  booktitle = {International Conference on Learning Representations},
  year = {2023}
}

@inproceedings{liu2023flowstraight,
  author = {Liu, Xingchao and Gong, Chengyue and Liu, Qiang},
  title = {Flow Straight and Fast: Learning to Generate and Transfer Data with Rectified Flow},
  booktitle = {International Conference on Learning Representations},
  year = {2023}
}

@inproceedings{poole2022dreamfusion,
  author = {Poole, Ben and Jain, Ajay and Barron, Jonathan T. and Mildenhall, Ben},
  title = {DreamFusion: Text-to-3D Using 2D Diffusion},
  booktitle = {International Conference on Learning Representations},
  year = {2023}
}

@inproceedings{lin2023magic3d,
  author = {Lin, Chen-Hsuan and Gao, Jun and Tang, Luming and Takikawa, Towaki and Zeng, Xiaohui and Huang, Xun and Kreis, Karsten and Fidler, Sanja and Liu, Ming-Yu and Lin, Tsung-Yi},
  title = {Magic3D: High-Resolution Text-to-3D Content Creation},
  booktitle = {Proceedings of the IEEE/CVF Conference on Computer Vision and Pattern Recognition},
  year = {2023}
}

@inproceedings{wang2023prolificdreamer,
  author = {Wang, Zhengyi and Lu, Cheng and Wang, Yikai and Bao, Fan and Li, Chongxuan and Su, Hang and Zhu, Jun},
  title = {ProlificDreamer: High-Fidelity and Diverse Text-to-3D Generation with Variational Score Distillation},
  booktitle = {Advances in Neural Information Processing Systems},
  year = {2023}
}

@inproceedings{chen2024gsgen,
  author = {Chen, Zilong and Wang, Feng and Wang, Yikai and Liu, Huaping},
  title = {Text-to-3D Using Gaussian Splatting},
  booktitle = {Proceedings of the IEEE/CVF Conference on Computer Vision and Pattern Recognition},
  year = {2024}
}

@inproceedings{tang2024dreamgaussian,
  author = {Tang, Jiaxiang and Ren, Jiawei and Zhou, Hang and Liu, Ziwei and Zeng, Gang},
  title = {DreamGaussian: Generative Gaussian Splatting for Efficient 3D Content Creation},
  booktitle = {International Conference on Learning Representations},
  year = {2024}
}

@inproceedings{yi2024gaussiandreamer,
  author = {Yi, Taoran and Fang, Jiemin and Wu, Guanjun and Xie, Lingxi and Zhang, Xiaopeng and Liu, Wenyu and Tian, Qi and Wang, Xinggang},
  title = {GaussianDreamer: Fast Generation from Text to 3D Gaussians by Bridging 2D and 3D Diffusion Models},
  booktitle = {Proceedings of the IEEE/CVF Conference on Computer Vision and Pattern Recognition},
  year = {2024}
}

@inproceedings{mildenhall2020nerf,
  author = {Mildenhall, Ben and Srinivasan, Pratul P. and Tancik, Matthew and Barron, Jonathan T. and Ramamoorthi, Ravi and Ng, Ren},
  title = {NeRF: Representing Scenes as Neural Radiance Fields for View Synthesis},
  booktitle = {Proceedings of the European Conference on Computer Vision},
  year = {2020}
}

@article{muller2022instant,
  author = {M{\"u}ller, Thomas and Evans, Alex and Schied, Christoph and Keller, Alexander},
  title = {Instant Neural Graphics Primitives with a Multiresolution Hash Encoding},
  journal = {ACM Transactions on Graphics},
  volume = {41},
  number = {4},
  year = {2022}
}

@article{lombardi2019neuralvolumes,
  author = {Lombardi, Stephen and Simon, Tomas and Saragih, Jason and Schwartz, Gabriel and Lehrmann, Andreas and Sheikh, Yaser},
  title = {Neural Volumes: Learning Dynamic Renderable Volumes from Images},
  journal = {ACM Transactions on Graphics},
  volume = {38},
  number = {4},
  year = {2019}
}

@article{lombardi2021mixture,
  author = {Lombardi, Stephen and Simon, Tomas and Schwartz, Gabriel and Zollhoefer, Michael and Sheikh, Yaser and Saragih, Jason},
  title = {Mixture of Volumetric Primitives for Efficient Neural Rendering},
  journal = {ACM Transactions on Graphics},
  volume = {40},
  number = {4},
  year = {2021}
}

@article{reiser2023merf,
  author = {Reiser, Christian and Szeliski, Richard and Verbin, Dor and Srinivasan, Pratul P. and Mildenhall, Ben and Geiger, Andreas and Barron, Jonathan T. and Hedman, Peter},
  title = {MERF: Memory-Efficient Radiance Fields for Real-Time View Synthesis in Unbounded Scenes},
  journal = {ACM Transactions on Graphics},
  volume = {42},
  number = {4},
  year = {2023}
}

@article{thies2019deferred,
  author = {Thies, Justus and Zollh{\"o}fer, Michael and Nie{\ss}ner, Matthias},
  title = {Deferred Neural Rendering: Image Synthesis Using Neural Textures},
  journal = {ACM Transactions on Graphics},
  volume = {38},
  number = {4},
  year = {2019}
}

@article{zwicker2002ewa,
  author = {Zwicker, Matthias and Pfister, Hanspeter and van Baar, Jeroen and Gross, Markus},
  title = {EWA Splatting},
  journal = {IEEE Transactions on Visualization and Computer Graphics},
  volume = {8},
  number = {3},
  pages = {223--238},
  year = {2002}
}

@article{tewari2022advances,
  author = {Tewari, Ayush and Fried, Ohad and Thies, Justus and Sitzmann, Vincent and Lombardi, Stephen and Sunkavalli, Kalyan and Martin-Brualla, Ricardo and Simon, Tomas and Saragih, Jason and Nie{\ss}ner, Matthias and Pandey, Rohit and Fanello, Sean and Wetzstein, Gordon and Zhu, Jun-Yan and Theobalt, Christian and Agrawala, Maneesh and Shechtman, Eli and Goldman, Dan B. and Zollh{\"o}fer, Michael},
  title = {Advances in Neural Rendering},
  journal = {Computer Graphics Forum},
  volume = {41},
  number = {2},
  pages = {703--735},
  year = {2022}
}

@article{xie2022neuralfields,
  author = {Xie, Yiheng and Takikawa, Towaki and Saito, Shunsuke and Litany, Or and Yan, Shiqin and Khan, Numair and Tombari, Federico and Tompkin, James and Sitzmann, Vincent and Sridhar, Srinath},
  title = {Neural Fields in Visual Computing and Beyond},
  journal = {Computer Graphics Forum},
  volume = {41},
  number = {2},
  pages = {641--676},
  year = {2022}
}

@article{berger2017survey,
  author = {Berger, Matthew and Tagliasacchi, Andrea and Seversky, Lee M. and Alliez, Pierre and Guennebaud, Ga{\"e}l and Levine, Joshua A. and Sharf, Andrei and Silva, Claudio T.},
  title = {A Survey of Surface Reconstruction from Point Clouds},
  journal = {Computer Graphics Forum},
  volume = {36},
  number = {1},
  pages = {301--329},
  year = {2017}
}

@article{peyre2019computational,
  author = {Peyr{\'e}, Gabriel and Cuturi, Marco},
  title = {Computational Optimal Transport},
  journal = {Foundations and Trends in Machine Learning},
  volume = {11},
  number = {5--6},
  pages = {355--607},
  year = {2019}
}

@article{rubner2000emd,
  author = {Rubner, Yossi and Tomasi, Carlo and Guibas, Leonidas J.},
  title = {The Earth Mover's Distance as a Metric for Image Retrieval},
  journal = {International Journal of Computer Vision},
  volume = {40},
  number = {2},
  pages = {99--121},
  year = {2000}
}

@article{besl1992method,
  author = {Besl, Paul J. and McKay, Neil D.},
  title = {A Method for Registration of 3-D Shapes},
  journal = {IEEE Transactions on Pattern Analysis and Machine Intelligence},
  volume = {14},
  number = {2},
  pages = {239--256},
  year = {1992}
}

@article{wang2004ssim,
  author = {Wang, Zhou and Bovik, Alan C. and Sheikh, Hamid R. and Simoncelli, Eero P.},
  title = {Image Quality Assessment: From Error Visibility to Structural Similarity},
  journal = {IEEE Transactions on Image Processing},
  volume = {13},
  number = {4},
  pages = {600--612},
  year = {2004}
}

@article{fu20213dfuture,
  author = {Fu, Huan and Cai, Bowen and Gao, Lin and Zhang, Ling-Xiao and Wang, Cao and Li, Zengqi and Xun, Chengyue and Sun, Chengyue and Fei, Yiyun and Zheng, Yu and Li, Ying and Liu, Yi and Liu, Peng and Ma, Lin and Weng, Le and Hu, Xiaohang and Ma, Xin and Qian, Qian and Jia, Rongfei and Zhao, Binqiang and Zhang, Hao},
  title = {3D-FUTURE: 3D Furniture Shape with Texture},
  journal = {International Journal of Computer Vision},
  volume = {129},
  pages = {3313--3337},
  year = {2021}
}

@misc{chang2015shapenet,
  author = {Chang, Angel X. and Funkhouser, Thomas and Guibas, Leonidas and Hanrahan, Pat and Huang, Qixing and Li, Zimo and Savarese, Silvio and Savva, Manolis and Song, Shuran and Su, Hao and Xiao, Jianxiong and Yi, Li and Yu, Fisher},
  title = {ShapeNet: An Information-Rich 3D Model Repository},
  year = {2015},
  eprint = {1512.03012},
  archivePrefix = {arXiv}
}

@inproceedings{deitke2023objaverse,
  author = {Deitke, Matt and Schwenk, Dustin and Salvador, Jordi and Weihs, Luca and Michel, Oscar and VanderBilt, Eli and Schmidt, Ludwig and Ehsani, Kiana and Kembhavi, Aniruddha and Farhadi, Ali},
  title = {Objaverse: A Universe of Annotated 3D Objects},
  booktitle = {Proceedings of the IEEE/CVF Conference on Computer Vision and Pattern Recognition},
  year = {2023}
}

@inproceedings{collins2022abo,
  author = {Collins, Jasmine and Goel, Shubham and Deng, Kenan and Luthra, Achleshwar and Xu, Leon and Gundogdu, Erhan and Zhang, Xi and Vicente, Tomas F. Yago and Dideriksen, Thomas and Arora, Himanshu and Guillaumin, Matthieu and Malik, Jitendra},
  title = {ABO: Dataset and Benchmarks for Real-World 3D Object Understanding},
  booktitle = {Proceedings of the IEEE/CVF Conference on Computer Vision and Pattern Recognition},
  year = {2022}
}

@inproceedings{wu2023omniobject3d,
  author = {Wu, Tong and Zhang, Jiarui and Fu, Xiao and Wang, Yuxin and Ren, Jiawei and Pan, Liang and Wu, Wayne and Yang, Lei and Wang, Jiaqi and Qian, Chen and Lin, Dahua and Liu, Ziwei},
  title = {OmniObject3D: Large-Vocabulary 3D Object Dataset for Realistic Perception, Reconstruction and Generation},
  booktitle = {Proceedings of the IEEE/CVF Conference on Computer Vision and Pattern Recognition},
  year = {2023}
}

\end{document}